\documentclass[]{article}
\usepackage[letterpaper]{geometry}
\usepackage{amta2022}
\usepackage{times}
\usepackage{url}
\usepackage{latexsym}
\usepackage{natbib}
\usepackage{layout}
\usepackage{xcolor}
\usepackage{lettrine}
\usepackage{graphicx}
\usepackage{tikz}

\usepackage{amsmath}
\usepackage{amsfonts}
\usepackage{amssymb}

\begin{document}

% \amtaHeader{x}{x}{xxx-xxx}{2015}{45-character paper description goes here}{Author(s) initials and last name go here}
\title{\bf Consistent Human Evaluation \\ of  Machine Translation across Language Pairs}

\author{\name{\bf Daniel Licht} \addr{META AI}
        \hfill \addr{dlicht@fb.com}
\AND
        \name{\bf Cynthia Gao} \addr{META AI}
        \hfill \addr{cynthiagao@fb.com}
\AND
        \name{\bf Janice Lam} \addr{META AI}
        \hfill \addr{janilam@fb.com}
\AND
        \name{\bf Francisco Guzm\'an} \addr{META AI}
        \hfill \addr{fguzman@fb.com}
\AND
        \name{\bf Mona Diab} \addr{META AI}
        \hfill \addr{mdiab@fb.com}
\AND
        \name{\bf Philipp Koehn} \addr{META AI / Johns Hopkins University}
        \hfill \addr{phi@jhu.edu}
}

\maketitle
\pagestyle{empty}

\begin{abstract}
Obtaining meaningful quality scores for machine translation systems through human evaluation remains a challenge given the high variability between human evaluators, partly due to subjective expectations for translation quality for different language pairs. We propose a new metric called XSTS that is more focused on semantic equivalence and a cross-lingual calibration method that enables more consistent assessment. We demonstrate the effectiveness of these novel contributions in large scale evaluation studies across up to 14 language pairs, with translation both into and out of English.
\end{abstract}

%\todo{note: max 10 pages + references}

\section{Introduction}
While machine translation systems are typically evaluated with automatic metrics like BLEU, the gold standard for quality assessment is  evaluation of machine translation output by human evaluators. Note that the validity of automatic metrics is justified by correlation to human evaluations.

However, human evaluators apply very different standards when assessing machine translation output, depending on their expectation of translation quality, their exposure to machine translation output, their language abilities, the presentation of source or reference translation, and vague category descriptions like "mostly correct". This is especially a problem when the goal is to obtain meaningful scores across language pairs, to assess, for instance, if a machine translation system for any given language pair is of sufficiently high quality to be put into use.

We address this problem of high variability and cross-lingual consistency by two novel contributions: (1) a novel scoring metric  XSTS that is focused on meaning and (2) an evaluation protocol that allows for calibration of scores across evaluators and across language pairs.
Our studies show that the XSTS score yields higher inter-annotator agreement compared against the more commonly used Direct Assessment \citep{graham-etal-2013-continuous}. We also show that our calibration leads to improved correlation of system scores to our subjective expectations of quality based on linguistic and resource aspects as well as improved correlation with automatic scores.

\section{Related Work}
The DARPA evaluation of the 1990s tasked human evaluators to assign scores from 1 to 5 to judge the fluency and adequacy of translations \citep{white-oconnell-1996-adaptation}, with vague definitions like {\em much meaning} for an adequacy score of 3 or the slightly offensive {\em non-native English} for a fluency score of 3. This scale was also used in the first human evaluation of the Workshop on Statistical Machine Translation (WMT) \citep{koehn-monz-2006-manual}. 

Note that these evaluations aim at a different goal than the one we are concerned with here: their main purpose is to rank the output of different machine translation systems against one another --- without the need to report a meaningful score that is an absolute measure of their translation quality. Hence, it should come as no surprise that the WMT evaluation then moved towards pairwise comparisons of different system outputs \citep{callison-burch-etal-2007-meta}. For many years, evaluators were asked to rank up to 5 system outputs against each other.

Due to the problem that for $n$ systems, $O(n^2)$ pairwise comparisons need to be done \citep{bojar-etal-2016-findings}, recent WMT evaluations switched to Direct Assessment \citep{graham-etal-2013-continuous}. Evaluators are required to indicate absolute quality of a machine translated sentence using a slider which is converted into a score on a 100 point scale. Such finer grained scores allow for easier normalization of scores between annotators. Direct Assessment is also used by Microsoft for shipping decisions \citep{kocmi-etal-2021-ship}. Google uses a 5-point scale to evaluate their machine translation systems but specifics have not been published.

Recently, \cite{mqm} proposed the Multidimensional Quality Metrics (MQM) Framework, rooted in the need for quality assurance for professional translators, that aims at generating meaningful scores. In MQM, fine-grained error categories like {\em omission, register} and {\em capitalization} are assessed and the error counts per category are combined into a single score. Such fine-grained errors can typically only be detected in relatively high-quality translations \citep{10.1162/tacl_a_00437}. This metric is predominantly used for quality assurance in the translation industry to evaluate translations from professional translations.

\section{A New Metric: XSTS}
We propose a new metric that is inspired by the Semantic Text Similarity metric (STS) used in research on paraphrase detection and textual entailment \citep{agirre-etal-2012-semeval}. The metric emphasises adequacy rather than fluency. We do this for several reasons but mainly because we deal with many low resource language pairs where preservation of meaning during translation is a pressing challenge. Arguably, assessing fluency is also much more subjective and thus leads to higher variance. Another reason is that we are interested in evaluating the translation of social media text where the source and reference translation may be disfluent, so lack of fluency should not be counted against machine translation. %really good point but also NMT typically has fluent output sometimes at the detriment of adequacy

As in many previously proposed scoring rubrics, we use a 5-point scale. For a detailed definition of the meaning of each score, see Figure~\ref{fig:xsts-definition}. There are various ways this metric could be used. The examples in the figure show two English sentences, such as machine translation output and a human reference translation, but our core evaluation protocol presents the source sentence and corresponding machine translation to a bilingual evaluator. Different from previous evaluation protocols, XSTS asks explicitly about meaning (semantic) correspondence, all the more while obfuscating which sentence is the source and which is the translation. %\todo{(See appendix for full annotation guidelines.)}

Note that the score has a fairly high bar for a score of 4: semantic equivalence, only allowing for differences in style, emphasis, and connotation. This allows us to detect differences in quality at the very high end. We experimented with both this 5 point scale and a reduced scale where the categories 4 and 5 were collapsed.

\begin{figure}\footnotesize
\lettrine[findent=2pt]{\fbox{\textbf{1}}}{ }
{\bf The two sentences are not equivalent, share very little details, and may be about different topics.  If the two sentences are about similar topics, but less than half of the core concepts mentioned are the same, then 1 is still the appropriate score.}
\vspace{1mm}

\noindent Example A (different topics):\\
\phantom{mm} Text 1: {\em John went horseback riding at dawn with a whole group of friends.}\\
\phantom{mm} Text 2: {\em Sunrise at dawn is a magnificent view to take in if you wake up early enough for it.}
\vspace{1mm}

\noindent Example B (similar/related topics):\\
\phantom{mm} Text 1: {\em  The woman is playing the violin.}\\
\phantom{mm} Text 2: {\em  The young lady enjoys listening to the guitar.}

\vspace{1mm}
\lettrine[findent=2pt]{\fbox{\textbf{2}}}{ } 
{\bf The two sentences share some details, but are not equivalent. Some important information related to the primary subject/verb/object differs or is missing, which alters the intent or meaning of the sentence.}
\vspace{1mm}

\hspace{-2mm}\begin{tabular}{p{6cm}p{6cm}}
Example A (opposite polarity): & Example B (word order changes meaning)\\
\phantom{mm} Text 1: {\em They flew out of the nest in groups.} & \phantom{mm} Text 1: {\em  James voted for Biden.}\\
\phantom{mm} Text 2: {\em They flew into the nest together.} & \phantom{mm} Text 2: {\em  Biden voted for James.}\\[1mm]
Example C (missing salient information): &  Example D (substitution/change in named entity)\\
\phantom{mm} Text 1: {\em ”He is not a suspect anymore.” John said.} & \phantom{mm} Text 1: {\em  I bought the book at Amazon.}\\
\phantom{mm} Text 2: {\em John said he is considered a witness but} & \phantom{mm} Text 2: {\em  The book was purchased at Barnes and}\\
\phantom{mmmmmm} {\em not a suspect.} & \phantom{mmmmmm} {\em Noble by me.}
\end{tabular}
\vspace{1mm}

\lettrine[findent=2pt]{\fbox{\textbf{3}}}{ } 
{\bf The two sentences are mostly equivalent, but some unimportant details can differ.  There cannot be any significant conflicts in intent or meaning between the sentences, no matter how long the sentences are.}
\vspace{1mm}

Example A (minor details that are not salient to the meaning):\\
\phantom{mm} Text 1: {\em  In May 2010, US troops invaded Kabul.}\\
\phantom{mm} Text 2: {\em  The US army invaded Kabul on May 7th last year, 2010.}
\vspace{1mm}

\noindent Example B (minor verb tense and/or unit of measurement differences):\\
\phantom{mm} Text 1: {\em  He bought 2 LBs of rice at Whole Foods.}\\
\phantom{mm} Text 2: {\em  He buy 1 KG. of rice at WholeFoods.}
\vspace{1mm}

\noindent Example C (small, non-conflicting differences in meaning):\\
\phantom{mm} Text1: {\em  She loves eating ripe apples in the fall.}\\
\phantom{mm} Text2: {\em  She usually eats ripened apple in autumn.}
\vspace{1mm}

\noindent Example D (omitted non-critical information, but no contradictory info introduced):\\
\phantom{mm} Text1: {\em  Several of the sailors set out on a rainy Tuesday morning.}\\
\phantom{mm} Text2: {\em  Several of the sailors set out on a Tuesday morning.}
\vspace{1mm}

\lettrine[findent=2pt]{\fbox{\textbf{4}}}{ } 
{\bf The two sentences are paraphrases of each other. Their meanings are near-equivalent,  with no major differences or information missing. There can only be minor differences in meaning due to differences in expression (e.g., formality level, style, emphasis, potential implication, idioms, common metaphors).}
\vspace{1mm}

\noindent Example A (different level of formality):\\
\phantom{mm} Text 1: {\em  This is Europe the so-called human rights country}\\
\phantom{mm} Text 2: {\em  This is Europe, the country of alleged human rights}
\vspace{1mm}

\noindent Example B (added sense of urgency, advertising style):\\
\phantom{mm} Text1: {\em  Special bike for more info call 0925279927}\\
\phantom{mm} Text2: {\em  Special bike for more information call now 0925279927}
\vspace{1mm}

\lettrine[findent=2pt]{\fbox{\textbf{5}}}{ } 
{\bf The two sentences are exactly and completely equivalent in meaning and usage expression (e.g., formality level, style, emphasis, potential implication, idioms, common metaphors).}
\vspace{1mm}

\hspace{-2mm}\begin{tabular}{p{6cm}p{6cm}}
Example A (same style and level of formality): & Example B (disfluency is not penalized):\\
\phantom{mm} Text 1: {\em  What’s up yu’all?} & \phantom{mm} Text 1: {\em  One two three apples oranges green}\\
\phantom{mm} Text 2: {\em  Howdy guys!} &\phantom{mm} Text 2: {\em  One two three apples oranges green}
\end{tabular}

    \caption{Part of the instruction given to evaluators to explain the XSTS scoring rubric. We also used a variant of this scale where 4 and 5 are collapsed into a single category.}
    \label{fig:xsts-definition}
\end{figure}

\section{Cross-Lingual Consistency via Calibration Sets}
Even after providing evaluators with instruction and training, they still show a large degree of variance in how they apply scores to actual examples of machine translation output. This is especially the case, when different language pairs are evaluated, which necessarily requires different evaluators assessing different output.

We address this problem with a calibration set. Note that we are either evaluating X--English or English--X machine translation systems. In either case, this requires evaluators who are fluent in English. Hence, we construct a calibration set by pairing machine translation output from various X--English systems with human reference translations --- so that the evaluators compare two English sentences. The sentence pairs are carefully chosen to cover the whole range of scores, based on consistent judgments from prior evaluation rounds.

Evaluators assess this fixed calibration set in addition to their actual task of assessing translations for their assigned language pair. We then compute the average score each evaluator gives to the calibration set. If this evaluator-specific calibration score is too high, then we conclude that the evaluator is generally too lenient and their scores for the actual task need to be adjusted downward, and vice versa.

There are various ways how scores for each evaluator could be adjusted. After exploring various options, we settled on a simple linear shift. To give an example, if the consensus score for the calibration set is 3.0 but an evaluator assigned it a score of 3.2, then we deduct 0.2 from all their scores for the actual evaluation task.

%\section{Trustworthy Human Evaluation of MT}

\section{Study Design}
We report on two large-scale human evaluation studies to assess the two novel contributions of this work. The first study compares XSTS and its variants against other evaluation methods like Direct Assessment. The second study assesses the effectiveness of our calibration method.

\paragraph{Language Pairs}
We selected languages with the goal to cover both high-resource languages with good machine translation quality and low-resource languages with weaker machine translation quality. The languages also differ in writing system, morphological complexity, and other linguistic dimensions. See the Table~\ref{tab:languages} for the list of languages in our studies.

\begin{table} \centering
\begin{tabular}{l|ll}
\bf Metric Study & \multicolumn{2}{l}{\bf Calibration Study}\\ \hline
Arabic & Amharic & Romanian \\  
Estonian & Arabic & Sindhi \\  
Indonesian & Azerbaijani & Slovenian\\  
Mongolian & Bosnian & Swahili \\  
Spanish & Georgian & Urdu \\
Tamil & Hindi & Zulu \\  
 & Brazilian Portuguese \\ \hline
\end{tabular}
\caption{Languages used. Both translation directions into and out of English were evaluated.}
\label{tab:languages}
\end{table}

\paragraph{Selection of Evaluators}
Evaluators were selected for each language pair and they evaluated both language directions (English--X and X--English). The evaluators were professional translators who were recruited by a translation agency. They had to have at least three years of translation experience, be native speakers of the language X, high level of English proficiency, and pass through a training process (detailed documentation of the task and training examples).
%\todo{The vetting process is the same as vetting for translators, these are the criteria
%• Professional translators
%• 3+ years’ translation experience in this language pair
%• Native speaker fluency in target language
%• High level in English (C2-C1)
%The training/onboarding process:
%Priority in terms of resourcing:
%1) existing FB approved resource pool
%2) existing vendor resource pool
%3) new onboarding
%Onboarding/training:
%• Profiling / CV screening
%• Signing the Framework agreement
%• Onboarding into Moravia production systems
%• Trainings – given the scope an standard onboarding time, the trainings are done in the form of providing detailed documentation, %instructions and training materials}

\paragraph{User Interface and Training} Since we are working with language service providers who subcontract the work to professional translators who differ in their technical setup, we do not always have full control over the way text is presented to them and how they register their evaluations. Throughout our studies, the employed tools vary from simple spreadsheets to a customized annotation tool similar to the one used in WMT evaluations.

\paragraph{Machine Translation Systems} Most of the machine translation systems used in this studies were trained in-house with fairseq \citep{ott-etal-2019-fairseq} on public data sets at different times in 2020 and 2021, each designed to optimized translation quality given available data and technology. The most recent system, used in the calibration study, is a 100-language multilingual system, similar to the one developed for the WMT 2021 Shared Task \citep{tran-etal-2021-facebook}.

\paragraph{Test Set} The translated sentences to be evaluated are selected from social media messages and Wikipedia --- the later being part of the FLORES test set which comprises close to 200 languages at the time of writing \citep{guzman-etal-2019-flores}. Note that social media messages have the additional challenge of disfluency and creative language variation in the source sentence.

\subsection{Study on Evaluation Metrics}
We compare the newly proposed XSTS to Direct Assessment and variants of XSTS. We report here on an experiment that used a 4-point XSTS scale but a subsequent study with a 5-point scale confirmed the findings. In all evaluations, the identity of the translation system was hidden and sentence translations of the different systems are randomly shuffled.

\begin{description}
\item[Direct Assessment (DA)]
In this protocol, the evaluators are required to judge translation output with respect to a source sentence on a 5-point qualitative rating scale. The evaluators render these ratings for  machine translations (MT1 , MT2, MT3) and a human translation (HT0), while shown the source sentence (source-based DA). 

\item[Cross Lingual Semantic Textual Similarity (XSTS)]
XSTS is the cross-lingual variant of STS. Evaluators indicate the level of correspondence between source and target directly. This protocol does not rely on reference translations. We apply XSTS to all directions for all translations (MT* and HT0). 

\item[Monolingual Semantic Textual Similarity (MSTS)]
MSTS is a protocol where the evaluators indicate the level of correspondence between two English strings, a machine translation (MT*) or human translation (HT0) and an additional human reference translation (HT1), using the XSTS scale.
This evaluation was only carried out for translations into English since we have two reference translations for English (HT0, HT1) but not for other languages.

\item[Back-translated Monolingual Semantic Textual Similarity (BT+MSTS)]
BT+MSTS is an attempt to make MSTS work for English-X translation when two reference translations are only available in English. Each translation from the English--X MT systems is manually back-translated into Englishf, which allows us to compare it against the English reference translations HT1 while also allowing for scoring the back-translation of HT0. Note that the manually back-translation will unlikely have fluency problems but any failures to preserve adequacy of the machine translation system will not be recovered by the professional translator.

\item[Post Editing with critical errors (PE)]
In this protocol, evaluators are required to provide the minimal necessary edits for the translations to render them correspondent to the source. Crucially however, evaluators are required to indicate the number of critical errors rendered in the post editing. The impetus behind this level of annotation is to transcend the traditional count of the number of edits needed to fix a translation. This protocol does not rely on a reference translation. Given the corrections, we computed three scores: critical edit counts, Levenshtein distance, and ChrF. 

\end{description}

%Pilot 4: As quality assurance of the evaluators, we use gold seeding and duplication. {\bf Gold seeding} uses carefully vetted examples that have a reliable score. They were obtain from prior evaluations where multiple evaluators agreed on the score \todo{any additional sanity check?}. 10\% of all evaluation examples are such gold examples, representing all possible scores. Evaluators can be assessed by how well they mirror these scores.  {\bf Duplication} means that we repeat evaluation examples after some time, and check if the same score is assign, allowing us to measure intra-annotator consistency. 20\% of evaluation examples are duplicated.

As test sets we used 250 sentences of social media messages (collected from public Facebook posts). We primarily report on results on this social media test set but an additional study on the Wikipedia test set (FLORES) confirms these findings.
We evaluated two internal machine translation systems (MT0 and MT1) and translations obtained from Google Translate (MT2). See Figure~\ref{tab:metrics-languages} for details on the languages involved.

\begin{table}[]
    \centering
    \begin{tabular}{l|c|c|c|c|c}
Language & Morphological & Resource & Writing & Inherent & Language \\
 & complexity & presence & system & variants & family\\ \hline
Arabic (AR) & xxxx & High & Arabic & Yes & Semitic \\
Estonian (ET) & xxx & Medium & Latin & No & Uralic\\
Indonesian (ID) & x & Medium & Latin & Yes & Austronesian\\
Mongolian (MN) & xxx & Low & Cyrillic & No & Mongolic\\
Tamil (TA) & xxx & Low & Tamil & No & Dravidian\\
Spanish (ES) & xx & High & Latin & Yes & Indo-European\\ \hline
    \end{tabular}
    \caption{Details on Languages used in the Metrics Study}
    \label{tab:metrics-languages}
\end{table}

%\todo{\url{https://docs.google.com/document/d/1J0Ona-gUMeW061ZmBXqMdj8c9gsSU3Cw49YtKwuHz0k/edit?pli=1#}}
%\todo{\url{https://docs.google.com/document/d/1ieSmiO6Qbmi974HXH1vToQLZubgOPewpOnN0rdZ_SPU/edit#}}

\subsection{Study on Calibration}
In a second study, we examined the introduction of a calibration set to create meaningful scores that can be compared across language pairs. This enables for instance the decision if a machine translation system for a language pair is good enough to be put into production.

Evaluators judge 1012 sentences pairs for a single language pair in both language directions. In this study, we only use the XSTS score. Translations are judged against the source sentence. Machine translations a generated with a state-of-the-art multilingual machine translation system. Evaluators also judge the human reference translation. 

The crucial addition to the sentence pairs to be judged is a calibration set of sentence pairs that is common across all languages. It consists of 1000 pairs of a machine translation into English and a corresponding English reference translation. These sentence pairs are carefully selected to span a wide quality range, based on human-scored translations from previous evaluations where multiple evaluators agreed on the score. These scores were used as the consensus quality score.

%\todo{More detail on this?}. \todo{Explain how the "official" score for the calibration is determined. - They are drawn from previous evaluated XSTS scored translations, paired with a reference translation, some of the '1' score items are also output of 'junk' m2m100 directions without scoring (because we didn't have enough scored '1's yet)}

A fair objection to using such a calibration set is that we are asking evaluators to perform two different tasks --- comparing machine translation against a source sentence (English and non-English), and comparing machine translation against a reference (English and English) --- but posit that they will use the same standards when making quality assessments. %\todo{But that's okay because we say so. - The idea is that their monolingual behavior will at least be indicative of their bilingual comparison behavior, though of course not perfectly so}

Because the calibration set is fixed, its quality is fixed, and the average score each evaluator assigns to the sentence pairs in the set should be the same.  Hence, we can use the actual score assigned by each evaluator and the official fixed score as the basis to make adjustments to each evaluator's score. For instance, if an evaluator gives the calibration too high score, then we detect that they are too lenient and their scores need to be corrected downward.

Note that there is also a second fixed point that could be used for score adjustment: the average score each evaluator gives to the reference translation. These professionally translated and vetted translations should receive high scores, and we could adjust each evaluator's scores so that they average adjusted score for reference translations is a fixed value. The underlying assumption here is that reference translations are of identical quality across all language pairs.

The calibration study generates a set of data points for each assigned score that contain the following information:
(1) language pair, (2) machine translation system, reference translation, or calibration set, (3) evaluator, (4) sentence pair, and (5) raw XSTS score.
%\begin{enumerate}\itemsep -3pt\vspace{-5pt}
%    \item language pair
%    \item machine translation system, reference translation, or calibration set
%    \item evaluator
%    \item sentence pair
%    \item raw XSTS score
%\end{enumerate}

So far, we discussed calibration to adjust the scores for each evaluator
%\todo{ actually the calibration is done at the language direction level, AFTER majority scores and an the overall quality metric is computed.  Doing it entirely at annotator level is not very compatible, as you would have to make an adjustment of say +0.2, which doesn't make sense if using majority scoring which we use because it is thought to be more robust, though it could be done at annotator level if we switched to doing a simple average over annotators}.
Our real goal, however, is to adjust scores for each language pair. Hence, we aggregate the individual data points into the following statistics: (1)  language pair, (2) machine translation system, reference translation, or calibration set. and (3) average of median raw XSTS scores. We first take judgments of different evaluators for the same translation and determine the median value. Then, we average these scores for each combination of language pair and translation source (machine translation, reference translation, calibration set).
%\begin{itemize}\itemsep -3pt\vspace{-5pt}
%    \item language pair
%    \item machine translation system, reference translation, or calibration set
%    \item 
%\end{itemize}

Based on this, we determine an adjustment function
\begin{equation}
    f_\text{language-pair}: \text{raw-score} \rightarrow \text{adjusted-score}
\end{equation}

The simplest form of this function is a linear shift $f(x) = x + \alpha$ where $\alpha$ is the adjustment parameter.
To ensure that adjusted scores agree on the consensus set, we compute $\alpha$ for each language pair as
\begin{equation}
\alpha_\text{language-pair} = \text{consensus-score} - \text{avg-median-score(language-pair,calibration-set)}
\end{equation}

With two fix points (score on calibration set and score on human reference translations), we use an adjustment formula $f(x) = \beta x + \alpha$ and determine the parameters $\alpha$ and $\beta$ in similar fashion.

\section{Results}

\subsection{Evaluation Metrics}
While automatic metrics are typically evaluated against gold standard human evaluation, we do not have such a gold standard when assessing different human evaluation protocols. Instead, we appeal to desirable aspects of human evaluation and assess these. Different evaluators should give the same translation the same score (inter-evaluator reliability). Evaluations should properly detect the quality difference between machine translation and gold standard human translation (meaningfulness). The amout of human effort for evaluations is also a significant factor (cost).

\paragraph{Inter-Evaluator Reliability}
Reliability measures the reproducibility of the measurements obtained during evaluation. Variability in ratings is an indication of complexity of the evaluation, lack of clarity in the guidelines rendering it highly subjective. It should be noted that evaluating translations is inherently subjective, yet protocols that are able to transcend the inherent subjectivity should yield more reproducible measures leading to more reliable protocols.

We use Fleiss Kappa to measure three-way inter-evaluator reliability scores.  Kappa numbers above 0.4 typically indicate moderate to excellent agreement (the higher the better). Table~\ref{tab:inter-evaluator} shows the average Kappa across all evaluators for all translations HT0, MT0, MT1, MT2 for each of the protocols. 

In Table 4, we also note the overall average Kappa per protocol across all languages (AVG). 
%We also report the standard deviation. Ideally, we would like for a winning protocol to illustrate the lowest variance and standard deviation. Hence we report 1-STDev. Finally, to decide the rank of the various protocols, we combine the Avg and 1-STDev \todo{(define! - I’m fairly sure that calculation was part of Anh’s statistics pipeline which was the source of Mona's IAA numbers for that pilot.  That said, I believe it was intended to be the mean variation between annotators per item.   mean over items of the (std over annotators).  1-STDev is better when closer to 1.)} as a final score per protocol. 
MSTS is the highest scoring protocol as it exhibits the highest average Kappa across languages per protocol. BT+MSTS also performs well. For the protocols that apply to both language directions XSTS (0.43 and 0.67, average 0.55) ranks above DA (0.34 and 0.52, average 0.43) and PE (0.31 and 0.53, average 0.42).   

%\todo{(Daniel:  So, I was highly dubious of these metrics after pilot 3 completed and I still am.  BT+MSTS, and MSTS got artificially good scores and a high level of agreement because the large majority of scores came back with the same value of "3".  To the point it was almost useless as an indicator of quality which blocked it's adoption (this showed up strongly in the S/N estimates).  The signal was very consistently washed out to a "3", so these IAA metrics score well, but I do NOT believe the reason for the high IAA was that the protocol working well, it was a numerical side-effect of so many of the scores coming back with the same rating, and not with a whole range of ratings 1-4 or 1-5 like DA and XSTS were.  You note this 2 paragraphs down.  To help with this problem we adopted Krippendorf's Alpha as an IAA metric to help compensate for these effects after that pilot.  IF we are going to include this table, etc. then I think this issue needs to be made clear. )} 

\begin{table}
    \centering
    \begin{tabular}{l|c|c|c|c|c|c|c|c}
         \multicolumn{9}{c}{\bf X--English}\\ \hline
& AR & ES & ET & ID & MN & TA & AVG & Rank\\ \hline
DA & 0.36 & 0.39 & 0.26 & 0.50 & 0.16 & 0.40 & 0.34  & 3\\
PE & 0.32 & 0.63 & 0.54 & 0.09 & 0.11 & 0.15 & 0.31  & 4\\
XSTS &0.64 & 0.29 & 0.50 & 0.12 & 0.19 & 0.81 & 0.43 & 2\\
MSTS & 0.57 & 0.48 & 0.46 & 0.56 & 0.52 & 0.62 & 0.54  & 1\\ \hline
\multicolumn{9}{c}{\bf English--X}\\ \hline
%& EN-AR & EN-ES & EN-ET &EN-ID & EN-MN & EN-TA & AVG & 1-STDev & Score \bf Rank\\ \hline
DA & 0.50 &0.63 & 0.67 & 0.35 & 0.23 & 0.74 & 0.52 &  4\\
PE & 0.54 & 0.56 & 0.56 & 0.45 & 0.37 & 0.68 & 0.53 &  3\\
XSTS & 0.85 & 0.60 & 0.49 & 0.99 & 0.57 & 0.50 & 0.67 &  1\\
BT+MSTS & 0.46 &0.43 & 0.47 & 0.43 & 0.48 & 0.46 & 0.46 &  2\\ \hline
\end{tabular}
    \caption{{\bf Fleiss Kappa Inter-Evaluator Reliability} averaged across HT0, MT0, MT1, and MT2 per protocol. AVG indicates average Kappa across all language directions per protocol. Rank is based on AVG, the highest rank (1) is a reflection of the protocol that yielded the inter-annotator agreement.}
    \label{tab:inter-evaluator}
\end{table}

\paragraph{Score difference between human reference translation and machine translation}
A simple test of the meaningfulness of each protocol is whether we can clearly see a distinction between Human level translation quality (manually yielded by humans) and our Machine Translation quality.  If a protocol cannot meaningfully distinguish between HT and MT then it will not be very useful as a quality measure. This measure makes two crucial assumptions: (1) human translation is indeed excellent and (2) the data selected for annotation evaluation is reflective of various levels of quality for the machine translation. Accordingly, both within each language, and overall between languages, we expect to see a clear progression of: {\em human translation (HT0) $>$ better machine translation (MT1) $>$ worse machine translation (MT2)}.

DA, XSTS, and PE passed this test and were reasonably good at separating the three types of translation.  But, at least with our sample sizes MSTS and BT-MSTS had a very difficult time distinguishing between HT0 and MT1 or MT1 and MT2, even in cases where the other protocols did not have that difficulty.

\subsection{Calibration}
The goal of calibration is the adjust raw human evaluation scores so that they reflect meaningful assessment the quality of the machine translation system for a given language pair.
When comparing different adjustment methods, we are faced with the problem, that there is no real ground truth. However, we do have some intuitions under which circumstances our machine translation systems will likely do well. More training data, the more related languages are to English in terms of proximity in the language family tree, low degree of syntactic and semantic divergence, or the same writing system should be correlated with better machine translation quality, in both English--X and X--English translation systems \citep{birch-etal-2008-predicting}.
See Table~\ref{tab:language-details-calibration} for such details for the languages used in the study. Note that the training data comes of sources of varying quality. For instance, the bulk of the Romanian data comes from Open Subtitles --- a notoriously noisy corpus.

\begin{table}[]
    \centering \small
    \begin{tabular}{l|c|c|c|c|r}
    \bf Language & \bf Language family & \bf Writing system & \multicolumn{2}{c|}{\bf Linguistic properties} & \bf Training size  \\ \hline
%    \bf Language & \bf Language & \bf Writing  & \multicolumn{2}{c|}{\bf Linguistic} & \bf Training   \\ 
%      & \bf   family & \bf system & \multicolumn{2}{c|}{\bf properties} & \bf corpus size  \\ \hline
    Amharic & Semitic & Ethiopian & SOV & fusional & 2,011,822 \\
    Arabic & Semitic & Arabic & VSO & fusional & 51,979,453 \\
    Bosnian & Balto-Slavic & Latin & SVO & fusional & 14,325,281 \\
    Bulgarian & Balto-Slavic & Cyrillic & SVO & fusional & 51,044,962 \\
    Georgian & Georgian-Zan & Georgian & SOV & agglutinative & 950,086 \\
    Hindi & Indo-Iranian & Devanagari & SOV & fusional & 8,607,078\\
    North Azerbaijani & Southern Turkic & Latin & SOV & agglutinative & 869,224 \\
    Portuguese (Brazil) & Italic & Latin & SVO & fusional & 57,210,510 \\
    Romanian & Italic & Latin & SVO & fusional & 63,737,708 \\
    Sindhi & Indo-Iranian & Arabic & SOV & fusional & 420,354 \\
    Slovenian & Balto-Slavic & Latin & SVO & fusional & 30,300,765 \\
    Swahili & Volta-Congo & Latin & SVO & agglutinative & 5,529,619 \\
    Urdu & Indo-Iranian & Arabic &  SOV &  fusional & 4,767,174 \\
    Zulu & Volta-Congo & Latin & SVO & agglutinative & 4,117,686 \\ \hline
    \end{tabular}
    \caption{Languages used in the calibration study and their properties. Training corpus size is number of sentence pairs of the publicly available parallel data used for training. Note that sometimes a large part of the training corpus is of low quality.}
    \label{tab:language-details-calibration}
\end{table}

Additionally, we can also compute the correlation to automatic scores such as the BLEU score. Of course, BLEU scores are notoriously meaningless, for instance they are highly dependent on the literalness of the human reference translation and typically lower when translating into morphologically richer targeted languages. But note that we are using a test set that is shared across all languages. Thus, BLEU scores for translation systems from any language into English are scored against exactly the same English reference translation. 

Table~\ref{tab:results-calibration-ranking} shows how the scores for the language pairs were adjusted from the raw baseline scores by (1) the consensus calibration score, (2) fixing the human translation score to 4.687 (determined by averaging scores given to human translations across all language pairs), and (3) both. Intuitively, one of the easiest language is Portuguese due the large amounts of data and closeness to English. After adjusting with the calibration score Portuguese-English ranks above Hindi--English and Arabic--English. 

\definecolor{darkgreen}{rgb}{0.0, 0.5, 0.0}
\definecolor{darkblue}{rgb}{0.0, 0.0, 0.7}
\begin{table}
    \centering
\newcommand{\boxoptions}{\setlength{\fboxsep}{1.5pt}\setlength{\fboxrule}{0pt}}
\newcommand{\Portuguese}{\boxoptions\fcolorbox{red}{red}{\textcolor{white}{Portuguese}}}
\newcommand{\Hindi}{\boxoptions\fcolorbox{brown}{brown}{\textcolor{white}{Hindi}}}
\newcommand{\Arabic}{\boxoptions\fcolorbox{darkgreen}{darkgreen}{\textcolor{white}{Arabic}}}
\newcommand{\Bulgarian}{\boxoptions\fcolorbox{darkblue}{darkblue}{\textcolor{white}{Bulgarian}}}
\newcommand{\Swahili}{\boxoptions\fcolorbox{purple}{purple}{\textcolor{white}{Swahili}}}
\begin{tabular}{p{6.1cm}p{6.1cm}}
\tiny \tabcolsep 2pt
\begin{tabular}{l|l|l|l}
\multicolumn{4}{c}{\bf X--English} \\ \hline
\bf Raw & \bf CS & \bf HT & \bf CS+HT\\ \hline
4.94 \Hindi & 4.73 Bosnian & 4.64 \Hindi & 4.64 \Hindi\\
4.88 Slovenian & 4.65 \Portuguese & 4.57 Slovenian & 4.56 Slovenian\\
4.75 Bosnian & 4.64 \Hindi & 4.51 \Portuguese & 4.53 \Portuguese \\
4.62 \Arabic & 4.58 \Arabic & 4.49 Bosnian & 4.52 Bosnian\\
4.61 \Portuguese & 4.47 Slovenian & 4.40 \Arabic & 4.43 \Arabic\\
4.56 Sindhi & 4.34 Sindhi & 4.30 Sindhi & 4.31 Sindhi\\
3.98 \Swahili & 4.19 \Bulgarian & 4.19 \Bulgarian & 4.19 \Bulgarian\\
3.96 Urdu & 3.98 \Swahili & 4.14 Urdu & 4.03 Urdu\\
3.74 Romanian & 3.89 Romanian & 3.98 \Swahili & 3.98 \Swahili\\
3.70 \Bulgarian & 3.86 Urdu & 3.83 Romanian & 3.86 Romanian\\
2.90 Amharic & 2.98 Amharic & 3.12 Azerbaijani & 2.98 Amharic\\
2.80 Zulu & 2.91 Zulu & 3.00 Zulu & 2.91 Zulu\\
2.66 Azerbaijani & 2.91 Azerbaijani & 2.92 Amharic & 2.91 Azerbaijani\\
1.04 Georgian & 1.02 Georgian & 1.12 Georgian & 0.90 Georgian\\
\end{tabular}
&
\tiny \tabcolsep 2pt
\begin{tabular}{l|l|l|l}
\multicolumn{4}{c}{\bf English-X} \\ \hline
\bf Raw & \bf CS & \bf HT & \bf CS+HT\\ \hline
4.95 \Hindi & 4.88 Bosnian & 4.68 Bosnian & 4.68 Bosnian\\
4.93 Slovenian & 4.82 \Portuguese & 4.67 \Hindi & 4.67 \Hindi\\
4.90 Bosnian & 4.74 \Arabic & 4.65 \Portuguese & 4.65 \Portuguese\\
4.79 \Arabic & 4.66 \Hindi & 4.63 Slovenian & 4.62 Slovenian\\
4.78 \Portuguese & 4.56 \Bulgarian & 4.58 \Bulgarian & 4.58 \Bulgarian\\
4.41 \Swahili & 4.52 Slovenian & 4.56 \Arabic & 4.58 \Arabic\\
4.12 Romanian & 4.41 \Swahili & 4.38 \Swahili & 4.38 \Swahili\\
4.07 \Bulgarian & 4.27 Romanian & 4.26 Romanian & 4.26 Romanian\\
3.97 Urdu & 3.86 Urdu & 4.14 Urdu & 4.03 Urdu\\
3.62 Zulu & 3.73 Zulu & 3.76 Zulu & 3.75 Zulu\\
3.15 Amharic & 3.23 Amharic & 3.12 Amharic & 3.21 Amharic\\
2.99 Georgian & 2.96 Georgian & 3.07 Georgian & 2.96 Georgian\\
2.24 Azerbaijani & 2.50 Azerbaijani & 2.63 Azerbaijani & 2.46 Azerbaijani\\
2.16 Sindhi & 1.94 Sindhi & 1.89 Sindhi & 1.97 Sindhi\\
\end{tabular}
\end{tabular}
% grep togther cal_scores_for_XSTS_paper.txt  | perl -ne 'chop; @a=split(/\t/); print $a[17]."\t".$a[2]."\t".sprintf("%.2f",$a[10])."\t".$a[3]."\n";' | grep -v bottom | sort -r | perl make-table.perl
    \caption{Adjustment of average XSTS scores based on fixing the score on the calibration set (CS), the human reference translation (HS) or both (CS+HS), compared to unadjusted scores. The languages Hindi, Portuguese, Arabic, Bulgarian, and Swahili are highlighted. CS adjustment ranks them more closely to our expectations based on corpus size and language similarity.}
    \label{tab:results-calibration-ranking}
\end{table}

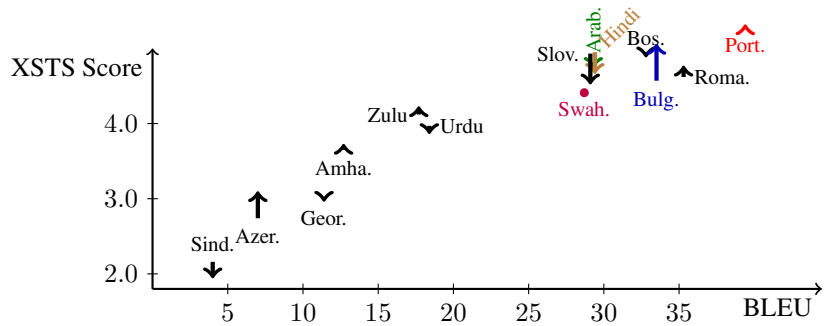
\begin{figure}
    \centering
        \begin{tikzpicture}
% draw x , y lines
\draw[thick,->] (0,0) -- (8.9,0) node[below left] {BLEU};
\draw[thick,->] (0,0) -- (0,3.2) node[below left] {XSTS Score};
% draw x ,y points (Values)
\foreach \i [count=\j from 0] in {5, 10, 15, 20, 25, 30, 35}
{
   \draw (\j+1,2pt) -- ++ (0,-4pt) node[below] {$\i$};
}
\foreach \i [count=\j from 0] in {2.0, 3.0, 4.0}
{
   \draw (2pt,\j+.2) -- ++ (-4pt,0) node[left]  {$\i$};
}

\draw[ultra thick,<-] (2.54,1.93) -- (2.54,1.85) node[anchor=north] {\footnotesize Amha.};
\draw[ultra thick,->,darkgreen] (5.86,2.99) node[anchor=west,rotate=90] {\footnotesize Arab.} -- (5.86,2.94);
\draw[ultra thick,<-] (1.40,1.30) -- (1.40,0.94) node[anchor=north] {\footnotesize Azer.};
\draw[ultra thick,->] (6.56,3.10) node[anchor=south] {\footnotesize Bos.} -- (6.56,3.08);
\draw[ultra thick,<-,darkblue] (6.70,3.26) -- (6.70,2.77) node[anchor=north] {\footnotesize Bulg.};
\draw[ultra thick,->] (2.28,1.19) node[anchor=north] {\footnotesize Geor.} -- (2.28,1.16);
\draw[ultra thick,->,brown] (5.88,3.15) node[anchor=west,rotate=45] {\footnotesize Hindi} -- (5.88,2.86);
\draw[ultra thick,<-,red] (7.88,3.52) -- (7.88,3.48) node[anchor=north] {\footnotesize Port.};
\draw[ultra thick,<-] (7.06,2.97) -- (7.06,2.82) node[anchor=west] {\footnotesize Roma.};
\draw[ultra thick,->] (0.80,0.36) node[anchor=south] {\footnotesize Sind.} -- (0.80,0.14);
\draw[ultra thick,->] (5.82,3.13) node[anchor=east] {\footnotesize Slov.} -- (5.82,2.72);
\draw[ultra thick,->] (3.68,2.17) node[anchor=west] {\footnotesize Urdu} -- (3.68,2.06);
\draw[ultra thick,<-] (3.54,2.43) -- (3.54,2.32) node[anchor=east] {\footnotesize Zulu};
\filldraw[purple] (5.74,2.61) circle (1.5pt) node[anchor=north] {\footnotesize Swah.};
    \end{tikzpicture}
\vspace{-3mm}
    \caption{Adjusted X--English scores using calibration set increases correlation with BLEU.}
    \label{fig:results-calibation-correlation}
\end{figure}

The second method to assess the effectiveness of our calibration methods is by computing correlation. We measure correlation with 3 different statistical methods: Pearson's R, $r^2$, and LinReg\footnote{The "LinReg" score was calculated as the mean r2 goodness of fit metric for training a simple sklearn linear regression model on spmBLEU scores and the calibrated XSTS scores using k-fold Cross-Validation with a 1:1 Train/test split, with the cross-validation split randomly bootstraped 5000 times.}.  Results are shown in Table~\ref{tab:results-calibation-correlation} and an illustration in Figure~\ref{fig:results-calibation-correlation}. Independent of the correlation method, or if we compute correlation into English, out of English, or both, the calibration method of adjusting the score based on the calibration set yields the highest correlation, clearly outperforming the baseline of unadjusted scores.

\begin{table}
    \centering
    \begin{tabular}{l|c|c|c|c|c|c|c|c|c}
        \bf Method & \multicolumn{3}{c|}{\bf Pearson's R} & \multicolumn{3}{c|}{\bf $r^2$} & \multicolumn{3}{c}{\bf LinReg} \\ \cline{2-10}
        & X-EN & EN-X & both & X-EN & EN-X & both & X-EN & EN-X & both \\ \hline
        baseline &     .897 &     .711 &     .797 &     .804 &     .506 &     .635 &     .715 &     .288 &     .566 \\
        CS       & \bf .946 & \bf .772 & \bf .854 & \bf .895 & \bf .595 & \bf .730 & \bf .833 & \bf .342 & \bf .650 \\
        HT       &     .926 &     .749 &     .834 &     .858 &     .561 &     .696 &     .767 &     .276 &     .604 \\
        CS+HT    &     .934 &     .757 &     .839 &     .874 &     .573 &     .704 &     .803 &     .283 &     .612 \\ \hline
    \end{tabular}
    \caption{Correlation of XSTS scores with spmBLEU scores fixing the score on the calibration set (CS), the human reference translation (HS) or both (CS+HS), compared to  raw scores.}
    \label{tab:results-calibation-correlation}
\end{table}

%\todo{\url{https://docs.google.com/document/d/16RGBUbI3iOtMWCbcTfFlZhNRKTYDVkGaB1RARVO_S44/edit}}

\section{Conclusion}
We introduced two novel contribution to the human evaluation of machine translation for multiple language pairs and validated their effectiveness in industrial-scale user studies: We proposed the scoring metric XSTS which is focused on meaning and introduced a calibration method that allows us to achieve meaningful scores that rank the quality of machine translation systems for different language pairs so that they match more closely with our intuition (plausibility) and automatic scores.

\newpage
\small
\bibliographystyle{apalike}
\bibliography{amta2022}

\end{document}